\documentclass[11pt]{article}
\usepackage{geometry}
\usepackage{coling2020}
\usepackage{times}
\usepackage{url}
\usepackage{latexsym}
\usepackage{microtype}
\usepackage[font={bf, footnotesize}, textfont=md]{caption}
\usepackage{graphicx}
\usepackage{subfigure}
\usepackage{epstopdf}
\usepackage{makecell}

\graphicspath{{figures/},{pics/}}

\colingfinalcopy 

\title{YNU-HPCC at SemEval-2020 Task 11: LSTM Network for Detection of Propaganda Techniques in News Articles}

\author{Jiaxu Dao, Jin Wang and Xuejie Zhang \\
School of Information Science and Engineering \\
Kunming, China \\
Yunnan University \\
  {\tt Contact:\{wangjin, xjzhang\}@ynu.edu.cn} 
  }

\date{}

\begin{document}
\maketitle
\begin{abstract}
  This paper summarizes our studies on propaganda detection techniques
for news articles in the SemEval-2020 task 11. This task is divided
into the SI and TC subtasks. We implemented the GloVe word
representation, the BERT pretraining model, and the LSTM model
architecture to accomplish this task. Our approach achieved good
results for both the SI and TC subtasks. The macro-$F_{1}$-$score$ for
the SI subtask is 0.406, and the micro-$F_{1}$-$score$ for the TC
subtask is 0.505. Our method significantly outperforms the
officially released baseline method, and the SI and TC subtasks
rank 17th and 22nd, respectively, for the test set. This paper also
compares the performances of different deep learning model
architectures, such as the Bi-LSTM, LSTM, BERT, and XGBoost
models, on the detection of news promotion techniques. The code of this paper is availabled at: \url{https://github.com/daojiaxu/semeval_11}.
\end{abstract}

\section{Introduction}
\label{intro}

%
%

\blfootnote{
    %
    %
    \hspace{-0.65cm}  
    This work is licensed under a Creative Commons Attribution 4.0 International License. License details: \url{http://creativecommons.org/licenses/by/4.0/}.
    %
    %
    %
    %
}

Propaganda techniques need to attach importance to arouse the emotions
of the receivers, sometimes even by temporarily bypassing the intellectual
defenses of the receivers \newcite{Pearlin1952} . Propaganda uses
psychological and rhetorical techniques to achieve its purpose. Such
techniques include using logical fallacies and appealing to the
emotions of the audience. Logical fallacies are usually hard to spot
since the argumentation, at first, might appear correct and
objective \cite{DaSanMartino2019} . However, careful analysis
shows that the conclusion cannot be drawn from the premise without
misusing logical rules. Another set of techniques uses emotional
language to induce the audience to agree with the speaker only
based on the emotional bond that is being created, provoking the
suspension of any rational analysis of the argumentation.

The traditional NLP task generally classifies and detects propaganda
techniques at the article level, which often fails to meet more
detailed requirements. This fact has also been confirmed by previous
iterations of the SemEval competition, where leading solutions used
convolutional neural networks (CNN), long short-term
memory (LSTM)  \cite{Baziotis2018}  and
transfer learning techniques \cite{Duppada2018} .
The main features of an article are extracted by using the feature
capture and pooling of the CNN model, but these methods can only be
used at the article level and are coarse-grained detection
methods. However, limited research has focused on text
classification \cite{Lewis1995,Song2010Hierarchical}.
 News articles have also been classified using the Bi-LSTM-CNN
model \cite{Li2018} .   However, there are often many propaganda
techniques in one article, and most of these techniques are efficient
for propaganda classification but lack the ability to detect categories
of propaganda techniques. Thus, they cannot achieve good results and are
less efficient in practice.

Now the difficulty is to detect propaganda techniques at the fine-grained
level. The SemEval-2020 Task 11, ``Detection of Propaganda Techniques in
News Articles'', is designed to promote research on this task. We used
the word embedded representation of the pretrained model and LSTM model
to detect the news article propaganda techniques at a fine-grained
level, and we also evaluate the performance among different neural
network models on this task.

The task consists of two subtasks.
\begin{enumerate}
	\item Span Identification (SI): Given a plain-text document, identify those specific fragments that contain at least one propaganda technique \cite{DaSanMartinoSemeval20task11} .  This is a binary sequence tagging task. We need to detect which fragments of the news article belong to the propaganda technique and mark the fragments with begin\_offset and end\_offset. The span ranges from begin\_offset to end\_offset-1.
	\item Technique Classification (TC): The purpose of this subtask is to identify the category of the propaganda technique. Given a text fragment identified as propaganda and its document context, identify the applied propaganda technique in the fragment. Since there are overlapping spans, formally, this is a multilabel multiclass classification problem. However, whenever a span is associated with multiple techniques, the input file will have multiple copies of such fragments; therefore, the problem can be treated as a multiclass classification problem. The techniques include Appeal\_to\_Authority, Appeal\_to\_fear-prejudice, Black-and-White\_Fallacy, and so on.
\end{enumerate}

The rest of the paper is organized as follows. Section 2 describes the details of the LSTM used in our system. Section 3 presents the experimental results. Conclusions and future works are described in Section 4.

\section{System Description}

We implemented LSTM model to accomplish this task. Meanwhile, the representations of input words are trained by using GloVe model \cite{pennington-etal-2014-glove}.

\subsection{Span Identification (SI)}

For the SI subtask requirements, we need to detect which fragments
of the news article utilized a propaganda technique. The SemEval
organizers provided us with 371 training sets. The data were plain
text files, and the SI task was to identify specific pieces that
contained at least one propaganda technique. To detect news article
propaganda techniques, we tested some deep learning model and
integration architectures \cite{Chen2016} . For the SI subtask, we
also experimented with GloVe-BiLSTM  \cite{Li2017,Luo2018,cross-huang-2016-incremental} ,
BERT-LSTM, GloVe-LSTM and BERT-BiLSTM \cite{Agirre2016,MacAvaney2018} .
As illustrated in Figure 1, our model includes an embedding layer, an LSTM
layer, a fully connected layer and an output layer. First, the embedding
layer represents every word using pretrained word embeddings. The LSTM
layer is implemented to obtain contextual information. The hidden vector
proceeded by each LSTM cell will be further fed into a dense layer with
the   \emph{sigmoid} activation function. Then, we can discriminate whether a word
is propaganda or not. Finally, we record the index of propaganda words and
recognize the propaganda fragments.


  Based on our experimental results, we can be concluded that the
  LSTM model with GloVe word embeddings obtained the best performance
  on this task. For the embeddings layer, we implemented GloVe to
  train the word embeddings \cite{Papagiannopoulou2018}.  The
  input tokens were obtained using the NLTK toolkit on the given
  articles. After the word embedding representation trained by GloVe
  is obtained, an LSTM layer is connected. In LSTM, recurrent cells
  are connected in a special way to avoid vanishing and exploding
  gradients, and the number of hidden nodes in the LSTM layer is set
  to 150. We find that the Bi-LSTM model \cite{Ma2016} does not
  perform well on this task. Next, the features captured by the LSTM
  layer are flattened and passed to the hidden dense layer, and the
  parameters of the dense layer are set to 8, which analyzes the
  interactions among the obtained vectors. The dropout rate of the
  dense layer is set to 0.2 to prevent model overfitting.


\subsection{Technique Classification (TC)}

The TC task is a multiclass classification task representing an
extension of the SI task, and the TC subtask seeks to classify
the various propaganda techniques identified by the SI
subtask. Such techniques include the use of logical fallacies
and appealing to the emotions of the audience. Logical fallacies
are usually hard to spot since the argumentation, at first, might
seem correct and objective. For the TC subtask, we obtain the
sentence representations via two ways. The first is feeding
pretrained word embeddings into the LSTM model and then we treat
the last hidden vector as the sentence representation. The second
way is using a pretrained language model, such as BERT, to directly
produce sentence representations. Meanwhile, we also compare the
effects of  \emph{softmax} and XGBoost  \cite{Mitchell2017} on
the classification task. Through comparing the experimental results, we
can conclude that the BERT-LSTM model can obtain good performance on the
TC subtask. The detailed analysis of the experimental results of the TC
subtask will be introduced in Section 3 of this paper.

We will introduce the selected model and parameter setting for the
experiment. The BERT pretrained language model has been proved
efficient in many NLP tasks, and the pretraining model used in
our experiment is BERT-Base. Therefore, we implement BERT to train
the word representations for the TC subtask. As illustrated in
Figure 2, we implement BERT to train the word representations obtained
through the bert-as-service library. The 768-dimensional word embeddings
trained by BERT are fed to the LSTM layer, and the number of hidden
nodes in the LSTM layer is set to 50. Same as the SI task, the features
captured by the LSTM layer are flattened and passed to the hidden dense
layer, and the number of parameters for the dense layer is set to 32. The
dropout rate of the dense layer is set to 0.2 \cite{Srivastava2014}.

\begin{figure}[t!]
  \centering
  \begin{minipage}[t]{0.5\textwidth}
  \centering
  \includegraphics[width=7.5cm,height=6.5cm]{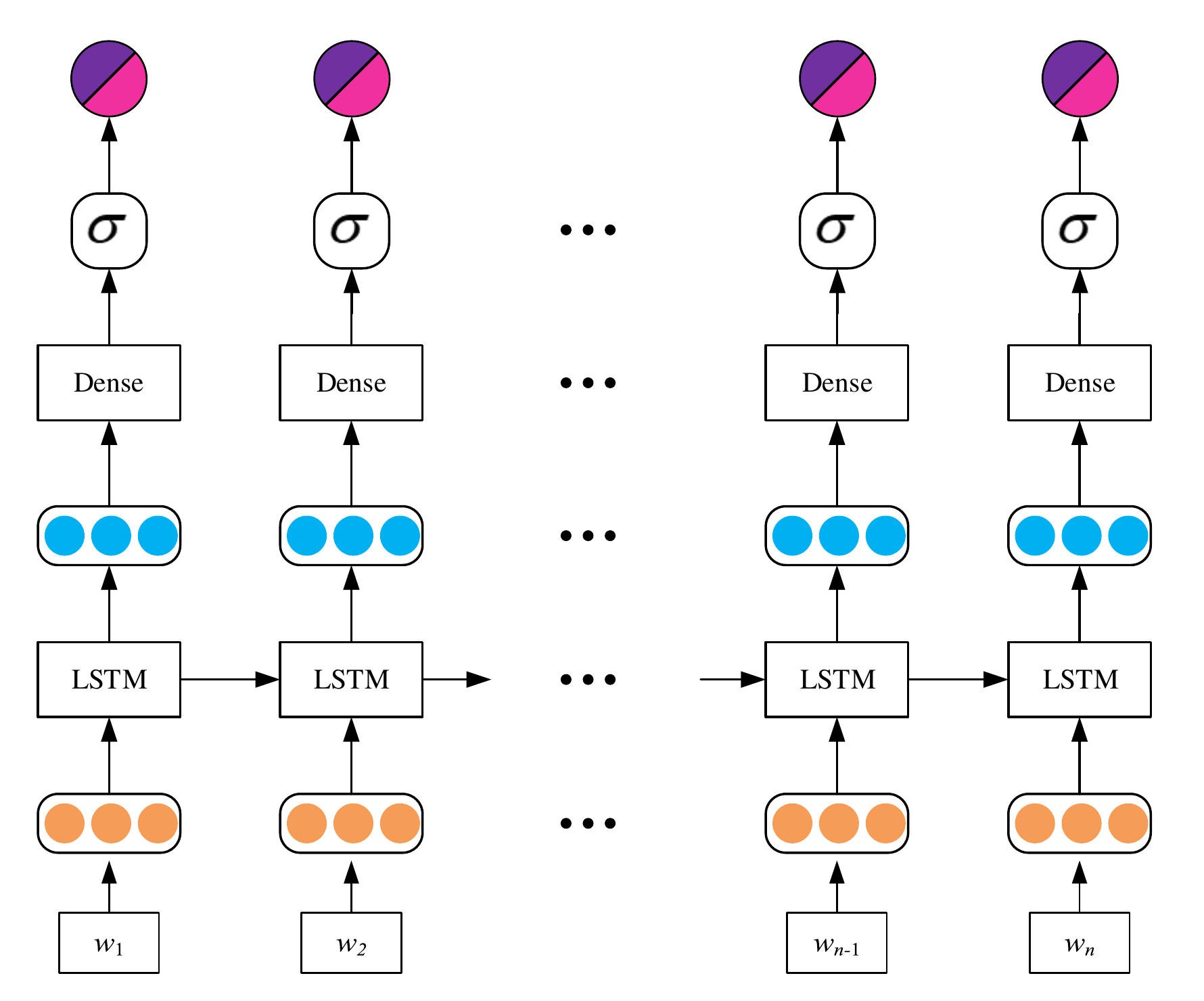}
  \caption{SI subtask model architecture.}
  \end{minipage}
  \begin{minipage}[t]{0.49\textwidth}
  \centering
  \includegraphics[width=7.5cm,height=8cm]{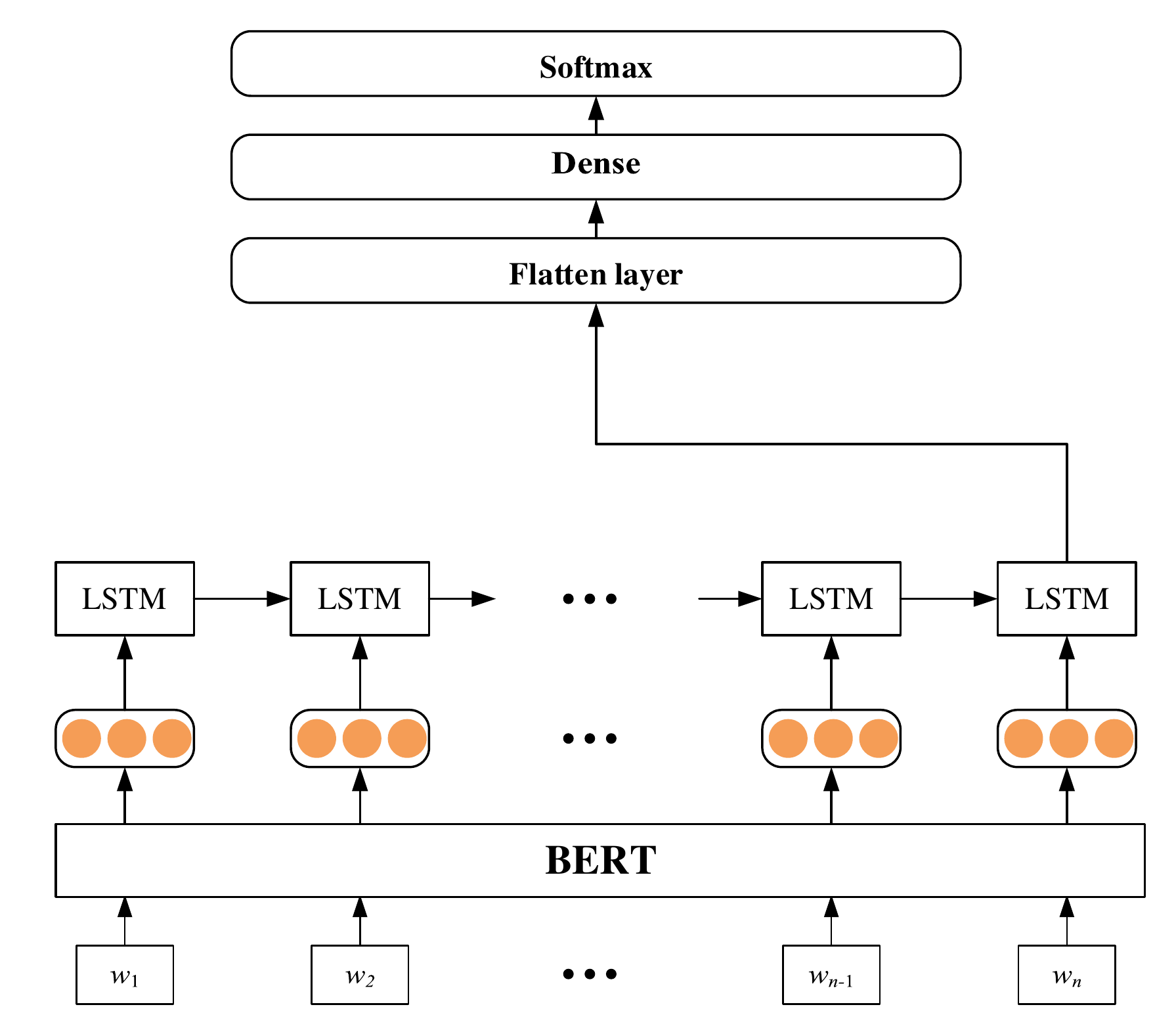}
  \caption{TC subtask model architecture.}
  \end{minipage}
  \end{figure}


\section{Experimental Results}
\label{sec:Results}

In Section 3, we first introduce the dataset of this task, and
then we analyze the performance of different neural networks and
integrated learning models for this task.

\subsection{Dataset}

The dataset contains 371 news articles for the training
set, 75 news articles for the development set, and 90 news
articles for the test set. The articles may contain several
propaganda spans. The beginning position and the ending position
were marked by ``begin\_offset'' and ``end\_offset'', respectively. As
illustrated in Table 1, for the ``111111111'' article, it contains 3
propaganda spans with a span range from ``begin\_offset'' to ``end\_offset'' minus
one because the index of words in the article started from zero.

\begin{table}[t!]
  \begin{center}
  \begin{tabular}{|c|c|c|p{4cm}<{\centering}|}
  \hline
  \small ID&\small Begin\_offset&\small End\_offset&\small Text \\
  \hline
  \small 111111111&\small 265&\small 323&\small The next transmission could be more pronounced or stronger \\
  \hline
  \small 111111111&\small 1069&\small 1091&\small a very, very different \\
  \hline
  \small111111111&\small1577&\small1616&\small but warned that the danger was not over \\
  \hline
\end{tabular}
\end{center}
\caption{ gold label SI file: article111111111.task1-SI. }
\end{table}

The TC subtask requires us to recognize the techniques of a
certain propaganda span. The propaganda technique and the
corresponding text contents are shown in Table 2. The number
of propaganda techniques is 18; therefore, the TC subtask is
a multiclass classification task.

\begin{table}[t!]
  \begin{center}
  \begin{tabular}{|c|c|c|c|p{4cm}<{\centering}|}
  \hline
  \small ID&\small Technique&\small Begin\_offset&\small End\_offset&\small Text \\
  \hline
  \small 111111111&\small Appeal\_to\_Authority&\small265&\small323&\small The next transmission could be more pronounced or stronger \\
  \hline
  \small111111111&\small Repetition&\small1069&\small1091&\small a very, very different  \\
  \hline
  \small111111111&\small Appeal\_to\_fear-prejudice&\small1577&\small1616&\small but warned that the danger was not over \\
  \hline
\end{tabular}
\end{center}
\caption{ gold label TC file: article111111111.task2-TC. }
\end{table}

\subsection{Evaluation Metrics}

For both subtasks, the participating systems were evaluated using
standard evaluation metrics, including the $accuracy$, $precision$,
$recall$ and $F_{1}$-$score$, which are calculated as follows:
\begin{equation}
  accuracy{\rm{ }} = \frac{{true{\rm{ }} \ positives + true{\rm{ }} \  negatives }}{{total{\rm{ }}\  \ number \ of \ instances}}
\end{equation}

\begin{equation}
  precision{\rm{ }} = {\rm{ }}\frac{{true{\rm{ }}\ positives}}{{true{\rm{ }}\ positives + {\rm{ }}false{\rm{ }}\ positives}}
\end{equation}

\begin{equation}
  recall{\rm{ }} = \frac{{true{\rm{ }}\ positives}}{{true{\rm{ }}\ positives + {\rm{ }}false{\rm{ }}\ negatives}}
\end{equation}

\begin{equation}
  {F_1}{\rm{ }} = 2 \cdot \frac{{precision \cdot recall}}{{precision + recall}}
\end{equation}

The organizers provided baseline models for each subtask. For the SI subtask, the macro-$F_{1}$-$score$ for the baseline model were 0.011 on the development set and 0.003 on the test set. For the TC subtask, the micro-$F_{1}$-$score$ for the baseline model was 0.265 on the development set and 0.252 on the test set.

\subsection{SI Subtask Results}

After fine-tuning the different parameters of the model, we finally
decided to use the adaptive moment estimation (Adam)
optimizer \cite{Kingma2015} with the learning rate set to 0.01. The
scores of the fine-tuning process of different learning rates on
the LSTM model with GloVe word embeddings are shown in Figure 3.

\begin{figure}[t!]
  \centering
  \includegraphics[
    height=6.72cm,width=8.94cm,
    keepaspectratio]
  {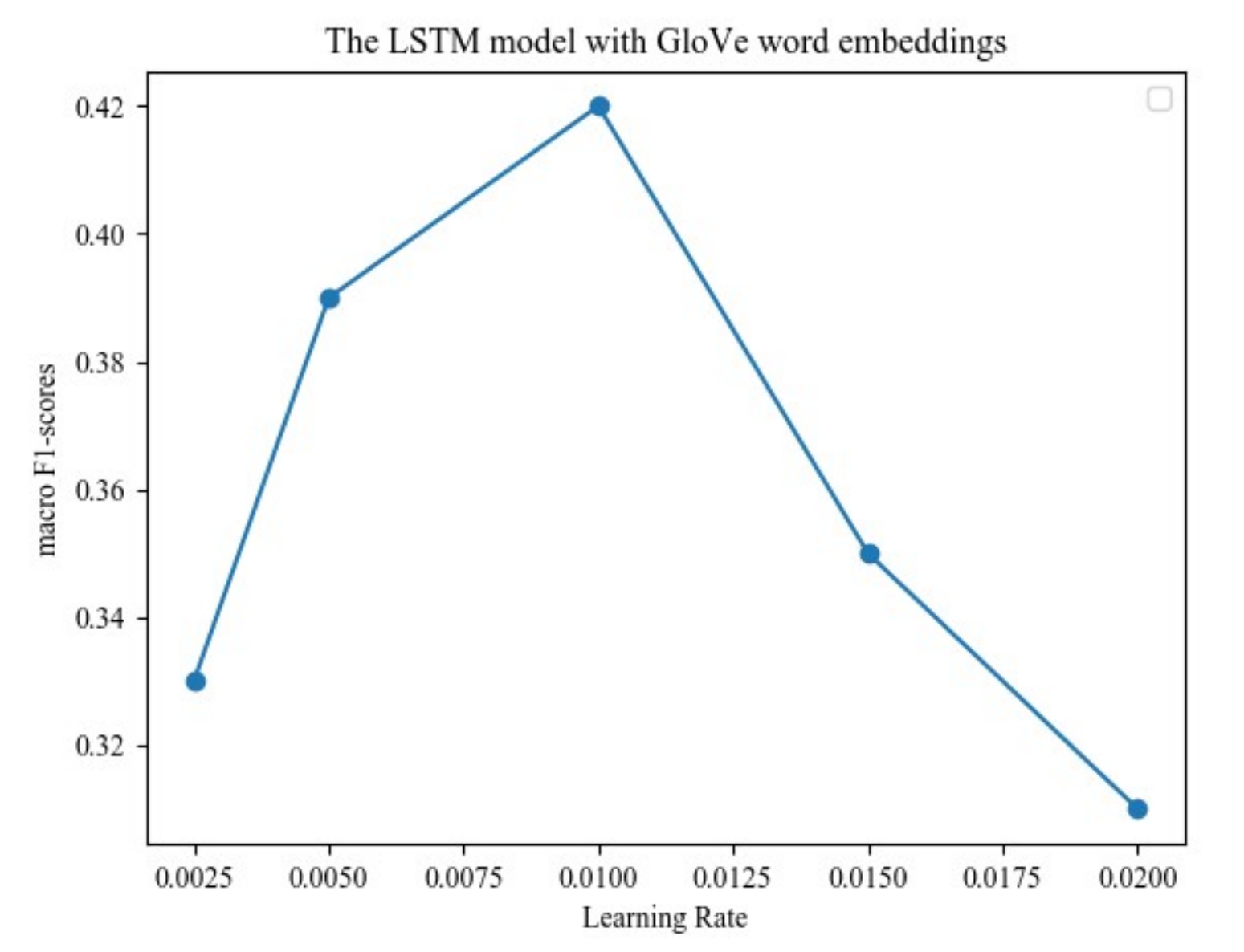}
  \caption{The fine-tuning process of different learning rates for the development set on the LSTM model with GloVe word embeddings.}
  \end{figure}

On the development set, the LSTM model with GloVe word embeddings
obtained a macro-$F_{1}$-$score$ of 0.423. In the test set, this
method achieved an $F_{1}$-$score$ of 0.406. The comparative results
are presented in Table 3.

\begin{table}[t!]
  \begin{center}
  \begin{tabular}{|c|c|c|p{1.5cm}<{\centering}|}
  \hline													
  \small \bf System&\small  \bf  $F_{1}$-$score$&\small  \bf $Precision$ &\small  \bf $Recall$ \\
  \hline
  \small GloVe+LSTM&\small  \bf 0.423&\small  \bf 0.321&\small  \bf 0.620	 \\
  \hline
  \small GloVe+BiLSTM&\small 0.404&\small 0.360&\small 0.460 \\
  \hline
  \small BERT+LSTM&\small 0.397&\small 0.287&\small 0.643 \\
  \hline
  \small BERT+BiLSTM&\small 0.360&\small 0.256&\small 0.608 \\
  \hline
\end{tabular}
\end{center}
\caption{  Scores of different models for the SI subtask on the development set. }
\end{table}

Our system ranked 17th out of 36 teams. The selected LSTM model with GloVe word embeddings significantly exceeded the system baseline in terms of performance, which proved that the model performed well on this task and could detect the span of propaganda techniques in news articles.

\subsection{TC Subtask Results}

TC is a multiclass classification task. As
illustrated in Table 4, the distribution of the
golden labels is rather imbalanced. Therefore, the
official evaluation measure for the task is the micro-$F_{1}$-$score$ .

\begin{table}[t!]
  \begin{center}
  \begin{tabular}{|c|p{5cm}<{\centering}|}
  \hline														
  \small \bf Propaganda Technique&\small  \bf Number Of TC Training Sets\\
  \hline
  \small APPEAL\_TO\_AUTHORITY&\small 155	 \\
  \hline
  \small APPEAL\_TO\_FEAR-PREJUDICE&\small 321 \\
  \hline
  \small BANDWAGON,REDUCTIO\_AD\_HITLERUM&\small 	77 \\
  \hline
  \small BLACK-AND-WHITE\_FALLACY&\small 	112 \\
  \hline
  \small CAUSAL\_OVERSIMPLIFICATION&\small 	212\\
  \hline
  \small DOUBT&\small 	517\\
  \hline
  \small EXAGGERATION,MINIMISATION&\small 	493\\
  \hline
  \small FLAG-WAVING	&\small 250\\
  \hline
  \small LOADED\_LANGUAGE&\small 	2200\\
  \hline
  \small NAME\_CALLING,LABELING&\small 	1105\\
  \hline
  \small REPETITION&\small 	621\\
  \hline
  \small SLOGANS&\small 	138\\
  \hline
  \small THOUGHT-TERMINATING\_CLICHES&\small 	80\\
  \hline
  \small WHATABOUTISM,STRAW\_MEN,RED\_HERRING&\small 	109\\
  \hline
\end{tabular}
\end{center}
\caption{The imbalanced data of the propaganda technique labels. }
\end{table}

The number of propaganda techniques is 18, but Table 4 only lists 14
techniques because some propaganda techniques are combined due to
insufficient data for some propaganda techniques in the
corpus \cite{DaSanMartinoSemeval20task11} . Since there are
overlapping spans, formally, it is a multilabel and multiclass
classification problem. However, whenever a span is associated
with multiple techniques, the input file will have multiple copies
of these fragments; therefore, the problem can be algorithmically
treated as a multiclass classification problem. We tried to use GloVe
and BERT to generate sentence embeddings, but the experimental results
showed that the sentence embedding produced by BERT pretraining was
better. After training on the datasets given by the TC
task  \cite{Drissi2019,Deriu2016} , through the model mentioned
above, the micro-$F_{1}$-$score$ on the development set was 0.561 and that on
the test set was 0.505. Our system ranked 22th out of 31 teams.

In addition to the LSTM model, we also tested some machine learning
architectures and some integrated learning methods. Because the
performance of the BERT-LSTM model on this task is better than those
of other models, we adopted the BERT-LSTM model as our final model. The
experimental results of the different models are shown in Table 5.

\begin{table}[t!]
  \begin{center}
  \begin{tabular}{|c|p{2cm}<{\centering}|}
  \hline													
  \small \bf System&\small  \bf $F_{1}$-$score$ \\
  \hline
  \small BERT+LSTM&\small  \bf 0.561	 \\
  \hline
  \small BERT+BiLSTM&\small 0.520 \\
  \hline
  \small BERT&\small 0.438 \\
  \hline
  \small BERT+XGBoost&\small 0.476 \\
  \hline
\end{tabular}
\end{center}
\caption{Scores of different models for the TC subtask on the development set. }
\end{table}

\section{Conclusion}

In this paper, we presented our system for the SemEval-2020 Task 11, which
leverages LSTM and pretrained word embeddings without using human-engineered
features for representation learning. Our experimental results show that the LSTM model with GloVe word
embeddings can get better performance according to the scores of different
neural network models and integration models on this task. The main goal
of this task is to detect propaganda techniques in news articles at a
fine-grained level, not just to make coarse judgments about whether the
news articles use propaganda techniques.\cite{Zhong2019ntuerAS}

It is known that neural networks perform well on large training sets, but sometimes a large, accurately labeled dataset cannot be obtained. For future work, the development of propaganda technology detection in news articles can be greatly improved in the pretraining model and the integrated model architecture.

\section*{Acknowledgements}
This work was supported by the National Natural Science Foundation of China (NSFC) under Grant No. 61966038, 61702443 and 61762091. The authors would like to thank the anonymous reviewers for their constructive comments.
\bibliographystyle{coling2020}
\bibliography{semeval2020}

\begin{thebibliography}{}

\bibitem[\protect\citename{Agirre \bgroup et al.\egroup }2016]{Agirre2016}
Eneko Agirre, Carmen Banea, Daniel Cer, Mona Diab, and Janyce Wiebe.
\newblock 2016.
\newblock {SemEval-2016 task 1: Semantic textual similarity, monolingual and
  cross-lingual evaluation}.
\newblock {\em SemEval 2016 - 10th International Workshop on Semantic
  Evaluation, Proceedings}, (7):497--511.

\bibitem[\protect\citename{Baziotis \bgroup et al.\egroup }2018]{Baziotis2018}
Christos Baziotis, Athanasiou Nikolaos, Pinelopi Papalampidi, Athanasia
  Kolovou, and Alexandros Potamianos.
\newblock 2018.
\newblock {NTUA-SLP at SemEval-2018 Task 3: Tracking Ironic Tweets using
  Ensembles of Word and Character Level Attentive RNNs}.
\newblock {\em Proceedings of The 12th International Workshop on Semantic
  Evaluation}, pages 613--621.

\bibitem[\protect\citename{Chen and Guestrin}2016]{Chen2016}
Tianqi Chen and Carlos Guestrin.
\newblock 2016.
\newblock {XGBoost: A scalable tree boosting system}.
\newblock {\em Proceedings of the ACM SIGKDD International Conference on
  Knowledge Discovery and Data Mining}, 13-17:785--794.

\bibitem[\protect\citename{Cross and Huang}2016]{cross-huang-2016-incremental}
James Cross and Liang Huang.
\newblock 2016.
\newblock Incremental parsing with minimal features using bi-directional
  {LSTM}.
\newblock In {\em Proceedings of the 54th Annual Meeting of the Association for
  Computational Linguistics (Volume 2: Short Papers)}, pages 32--37.

\bibitem[\protect\citename{Da~San~Martino \bgroup et al.\egroup
  }2020]{DaSanMartinoSemeval20task11}
Giovanni Da~San~Martino, Alberto Barr\'{o}n-Cede\~no, Henning Wachsmuth,
  Rostislav Petrov, and Preslav Nakov.
\newblock 2020.
\newblock {SemEval}-2020 task 11: Detection of propaganda techniques in news
  articles.
\newblock In {\em Proceedings of the 14th International Workshop on Semantic
  Evaluation}, SemEval 2020, Barcelona, Spain, December.

\bibitem[\protect\citename{Deriu \bgroup et al.\egroup }2016]{Deriu2016}
Jan Deriu, Aurelien Lucchi, Maurice Gonzenbach, Valeria~De Luca, Fatih Uzdilli,
  and Martin Jaggi.
\newblock 2016.
\newblock {SwissCheese at SemEval-2016 task 4: Sentiment classification using
  an ensemble of convolutional neural networks with distant supervision}.
\newblock {\em SemEval 2016 - 10th International Workshop on Semantic
  Evaluation, Proceedings}, pages 1124--1128.

\bibitem[\protect\citename{Drissi \bgroup et al.\egroup }2019]{Drissi2019}
Mehdi Drissi, Pedro {Sandoval Segura}, Vivaswat Ojha, and Julie Medero.
\newblock 2019.
\newblock {Harvey Mudd College at SemEval-2019 Task 4: The Clint Buchanan
  Hyperpartisan News Detector}.
\newblock {\em SemEval-2019}, pages 962--966.

\bibitem[\protect\citename{Duppada \bgroup et al.\egroup }2018]{Duppada2018}
Venkatesh Duppada, Royal Jain, and Sushant Hiray.
\newblock 2018.
\newblock {SeerNet at SemEval-2018 Task 1: Domain Adaptation for Affect in
  Tweets}.
\newblock {\em SemEval-2018}, pages 18--23.

\bibitem[\protect\citename{Kingma and Ba}2015]{Kingma2015}
Diederik~P. Kingma and Jimmy~Lei Ba.
\newblock 2015.
\newblock {Adam: A method for stochastic optimization}.
\newblock {\em 3rd International Conference on Learning Representations, ICLR
  2015 - Conference Track Proceedings}, pages 1--15.

\bibitem[\protect\citename{Lewis}1995]{Lewis1995}
David~D. Lewis.
\newblock 1995.
\newblock {Evaluating and optimizing autonomous text classification systems}.
\newblock {\em SIGIR Forum (ACM Special Interest Group on Information
  Retrieval)}, (1996):246--254.

\bibitem[\protect\citename{Li \bgroup et al.\egroup }2017]{Li2017}
Shudong Li, Zhou Yan, Xiaobo Wu, Aiping Li, and Bin Zhou.
\newblock 2017.
\newblock {A Method of Emotional Analysis of Movie Based on Convolution Neural
  Network and Bi-directional LSTM RNN}.
\newblock {\em Proceedings - 2017 IEEE 2nd International Conference on Data
  Science in Cyberspace, DSC 2017}, pages 156--161.

\bibitem[\protect\citename{Li \bgroup et al.\egroup }2018]{Li2018}
Chenbin Li, Guohua Zhan, and Zhihua Li.
\newblock 2018.
\newblock {News Text Classification Based on Improved Bi-LSTM-CNN}.
\newblock {\em Proceedings - 9th International Conference on Information
  Technology in Medicine and Education, ITME 2018}, pages 890--893.

\bibitem[\protect\citename{Luo \bgroup et al.\egroup }2018]{Luo2018}
Ling Luo, Zhihao Yang, Pei Yang, Yin Zhang, Lei Wang, Hongfei Lin, and Jian
  Wang.
\newblock 2018.
\newblock {An attention-based BiLSTM-CRF approach to document-level chemical
  named entity recognition}.
\newblock {\em Bioinformatics}, 34(8):1381--1388.

\bibitem[\protect\citename{Ma and Hovy}2016]{Ma2016}
Xuezhe Ma and Eduard Hovy.
\newblock 2016.
\newblock {End-to-end sequence labeling via bi-directional LSTM-CNNs-CRF}.
\newblock {\em 54th Annual Meeting of the Association for Computational
  Linguistics, ACL 2016 - Long Papers}, 2:1064--1074.

\bibitem[\protect\citename{MacAvaney \bgroup et al.\egroup
  }2018]{MacAvaney2018}
Sean MacAvaney, Luca Soldaini, Arman Cohan, and Nazli Goharian.
\newblock 2018.
\newblock {GU IRLAB at SemEval-2018 Task 7: Tree-LSTMs for Scientific Relation
  Classification}.
\newblock {\em SemEval-2018}, (2015):831--835.

\bibitem[\protect\citename{Martino \bgroup et al.\egroup
  }2019]{DaSanMartino2019}
Giovanni Da~San Martino, Seunghak Yu, Alberto Barrón-Cedeño, Rostislav
  Petrov, and Preslav Nakov.
\newblock 2019.
\newblock {Fine-Grained Analysis of Propaganda in News Article}.
\newblock {\em Proceedings of the 2019 Conference on Empirical Methods in
  Natural Language Processing and the 9th International Joint Conference on
  Natural Language Processing (EMNLP-IJCNLP)}, (4):5635--5645.

\bibitem[\protect\citename{Mitchell and Frank}2017]{Mitchell2017}
Rory Mitchell and Eibe Frank.
\newblock 2017.
\newblock {Accelerating the XGBoost algorithm using GPU computing}.
\newblock {\em PeerJ Computer Science}, 2017(7):27.

\bibitem[\protect\citename{Papagiannopoulou and
  Tsoumakas}2018]{Papagiannopoulou2018}
Eirini Papagiannopoulou and Grigorios Tsoumakas.
\newblock 2018.
\newblock {Local word vectors guiding keyphrase extraction}.
\newblock {\em Information Processing and Management}, 54(6):888--902.

\bibitem[\protect\citename{Pearlin and Rosenberg}1952]{Pearlin1952}
Leonard~I. Pearlin and Morris Rosenberg.
\newblock 1952.
\newblock {Propaganda Techniques in Institutional Advertising}.
\newblock {\em Public Opinion Quarterly}, 16(1):5.

\bibitem[\protect\citename{Pennington \bgroup et al.\egroup
  }2014]{pennington-etal-2014-glove}
Jeffrey Pennington, Richard Socher, and Christopher Manning.
\newblock 2014.
\newblock {G}lo{V}e: Global vectors for word representation.
\newblock In {\em Proceedings of the 2014 Conference on Empirical Methods in
  Natural Language Processing ({EMNLP})}, pages 1532--1543, Doha, Qatar,
  October. Association for Computational Linguistics.

\bibitem[\protect\citename{Song \bgroup et al.\egroup
  }2010]{Song2010Hierarchical}
Sheng~Li Song, Bao Liang, and Ping Chen.
\newblock 2010.
\newblock Hierarchical text classification and evaluation.
\newblock {\em Systems Engineering And Electronics}, 32(5):1088--1093.

\bibitem[\protect\citename{Srivastava \bgroup et al.\egroup
  }2014]{Srivastava2014}
Nitish Srivastava, Geoffrey Hinton, Alex Krizhevsky, Ilya Sutskever, and Ruslan
  Salakhutdinov.
\newblock 2014.
\newblock {Dropout: A simple way to prevent neural networks from overfitting}.
\newblock {\em Journal of Machine Learning Research}, 15:1929--1958.

\bibitem[\protect\citename{Zhong and Miao}2019]{Zhong2019ntuerAS}
Peixiang Zhong and Chunyan Miao.
\newblock 2019.
\newblock ntuer at {S}em{E}val-2019 task 3: Emotion classification with word
  and sentence representations in {RCNN}.
\newblock In {\em Proceedings of the 13th International Workshop on Semantic
  Evaluation}, pages 282--286.

\end{thebibliography}

\end{document}